\documentclass[runningheads]{llncs}
\usepackage[T1]{fontenc}
\usepackage{amsmath,amssymb,graphicx,url}
\usepackage{graphicx}
\usepackage{booktabs}
\usepackage{multirow}
\usepackage{float}
\usepackage{algorithm}
\usepackage[noend]{algorithmic}
\usepackage[numbers,sort&compress]{natbib}
\usepackage{arydshln}

\begin{document}
\graphicspath{ {./images/} } 
\title{Protected Probabilistic Classification Library}
\author{Ivan Petej}
\authorrunning{I. Petej}
\institute{Centre for Reliable Machine Learning, 
	Royal Holloway,  University of London\\
	Egham, Surrey TW20 0EX, UK}
%\email{i.petej@rhul.ac.uk}
%
\maketitle              % typeset the header of the contribution
\begin{abstract}
		This paper introduces a new \texttt{Python} package specifically designed to address calibration of probabilistic classifiers under dataset shift.  The method is demonstrated in binary and multi-class settings and its effectiveness is measured against a number of existing post-hoc calibration methods. The empirical results are promising and suggest that our technique can be helpful in a variety of settings for batch and online learning classification problems where the underlying data distribution changes between the training and test sets. 

\keywords{Machine learning classification, Probabilistic calibration, Online learning}
\end{abstract}

\section{Introduction}
	In machine learning, predictive models often output probability estimates that guide decision-making in high-stakes applications such as healthcare, finance, and autonomous systems. Such probability estimates must be well-calibrated, meaning that the predicted probabilities accurately reflect the true likelihood of events \cite{dawid1982well}. Poor calibration can lead to overconfident or under-confident predictions potentially causing erroneous decisions. Unfortunately, many standard probabilistic machine learning algorithms lack inherent calibration properties \cite{guo2017calibration}.\par
	Probabilistic calibration seeks to improve the reliability of model confidence scores by aligning predicted probabilities with actual observed frequencies. A number of post-hoc calibration approaches (i.e., methods applied after initial algorithm training), such as \emph{Platt scaling} \cite{platt1999probabilistic}, \emph{isotonic regression} \cite{zadrozny2002transforming}, \emph{binning} \cite{zadrozny2001learning}, \emph{Venn-ABERS} calibration~\cite{vovk2014venn} and \emph{beta} calibration \cite{kull2017beta} have been used to convert underlying classifier scores or probabilities into well-calibrated probabilities. Such methods have shown to perform well, especially when the underlying data in the test set follows the same distribution as the data in training and calibration sets \cite{silva2023classifier}.\par
In many real world applications however, as soon as the algorithm is trained, the data distribution changes and the prediction algorithm may need to be retrained or recalibrated. In this setting, it may be advantageous to utilize an algorithm which prevents a drop in the quality of calibration without the need to retrain the underlying classifier. In this paper we specifically address calibration of probabilistic classifiers under such dataset shift. Our work is an extension of the study originally reported in \cite{vovk2021protected}, extending it further to artificial and real-life multi-class problems. We further compare the  performance of our method with other recently reported post-hoc calibration techniques, including those specifically aimed at streaming data, with some promising results.

\section{Background and related work}

Good probabilistic calibration ensures that a model's predicted probabilities align with the true likelihood of outcomes, which is vital for applications involving risk-sensitive decisions. Over the years, research in this domain has addressed various aspects, including over and under-confidence of traditional machine learning algorithms, the loss of calibration under data distributional shift and the resulting design of a number of post-hoc calibration methods.\par
In addition to traditional post-hoc calibration methods described in Section 1,  \citet{guo2017calibration} were among the first to rigorously analyse the probabilistic calibration of deep neural networks. They observed that despite improvements in accuracy, deep neural network models tend to produce poorly calibrated probabilities. They proposed \emph{temperature scaling}, a simple yet effective post-processing technique to improve calibration without altering accuracy. In other work, ~\citet{ovadia2019can} evaluated the reliability of predictive uncertainty under dataset shift through large-scale empirical analysis. They demonstrated that while some models perform well under i.i.d. settings, they often fail to maintain calibrated uncertainty estimates under distributional shift, thereby highlighting the need for shift-resilient calibration methods.~\citet{li2020bayesian} proposed a Bayesian online learning approach for detecting and adapting to irregular distribution shifts. Their model combines Bayesian inference with change-point detection, enabling dynamic adaptation to evolving data distributions while maintaining probabilistic integrity.\par 
In the context of classification under label shift,~\citet{podkopaev2021distribution} developed a distribution-free uncertainty quantification framework using conformal prediction. This method provides prediction sets with finite-sample guarantees, without requiring strong distributional assumptions. In other work ~\citet{alvarez2021probabilistic} introduced a probabilistic load forecasting approach using adaptive online learning. Their framework updates forecasts in real time, achieving calibrated predictions in non-stationary environments. \citet{gupta2023online} extended calibration techniques to online settings by proposing \emph{Online Platt Scaling with Calibeating}. This method enables real-time updating of calibration parameters, outperforming static methods by adapting to streaming data. Finally, a recent survey by~\citet{minderer2024classifier} reviews the state of classifier calibration, covering both classical and modern techniques, metrics, and practical considerations. This study serves as a valuable resource for both theoreticians and practitioners looking to assess and improve calibration performance.\par
Together, these contributions represent am evolving body of work aimed at improving the reliability, adaptivity, and robustness of probabilistic predictions in machine learning systems to date. The aim of this work is to add and expand on a complementary method in the growing body of literature tackling machine learning algorithm adaptation under dataset shift.

\subsection{Main contributions}

The main contributions of this paper are:
\begin{itemize}
	\item  We introduce a new \texttt{Python} package for protected classification\footnote{available at \url{https://pypi.org/project/protected-classification}} and use it to extend a set of experimental results across a range of real and artificial datasets, in binary and multi-class settings
	\item We compare the method of protected classification in terms of calibration with a recently published alternative method \cite{gupta2023online}
	\item We demonstrate the effectiveness of protected classification in an online setting for streaming data problems  - to our knowledge this is the first such study to date
\end{itemize}

\section{Theoretical background}
Theoretical exposition of protected probabilistic classification has been extensively covered in \cite{vovk2021protected} and here we briefly summarizes the key results for context.

Our starting assumption is that we have access to a prediction model which is obtained by training an algorithm. The algorithm maps  a new object from the object space $\mathbf{X}$ to a label from the label space $\mathbf{Y}$ using past data of $x_1 \dots x_n \in \mathbf{X}$  and labels $y_1 \dots y_n \in \mathbf{Y}$. In the simple case of binary classification for example, such a system maps a given object $\mathbf{x}$ with an unknown label $y \in  [0,1]$  to a number $p  \in  [0,1]$, interpreted as the predicted probability that $y = 1$. We call such a system the \emph{base predictive system}.

We have no guarantees that the base predictive system is calibrated  - implying that for any given $P(y|x)  = p$ the average percentage of the true label of $Y$ is $p$.  To calibrate the base predictive system, we define a family $\mathbf{\Theta}$ of calibrating functions $f _{\theta} : [0,1] \rightarrow [0,1], \theta \in \mathbf{\Theta}$. The intuition behind any given calibration function  $f _{\theta}$ is that we are trying to improve the base predictions $p$, by using  a new prediction $f(p)$ instead of $p$. 

In our library, and in line with \cite{vovk2021protected} we make use of the Cox \cite{cox1958two} calibrating functions, which for a general multi-class case are defined in Equation \ref{eq:1} as:

\begin{equation}
	\label{eq:1}
	f_{\alpha, \beta}(\mathbf{p})_y := \frac{p_y^{\beta} \exp(\alpha(y))}{\sum_{y' \in \mathcal{Y}} p_{y'}^{\beta} \exp(\alpha(y'))},
\end{equation}

\noindent where the parameters $\alpha \in \mathbb{R}^{\mathbb{Y}}$,  $\beta \in \mathbb{R}$. We are mainly interested in values of $\beta \in \{1, 0.5, 2\}$ and $\alpha(y) \in \{0,  1,  -1\}$ for each class label $y$.

\begin{figure}[h]
	\centering
	\includegraphics[width=\textwidth]{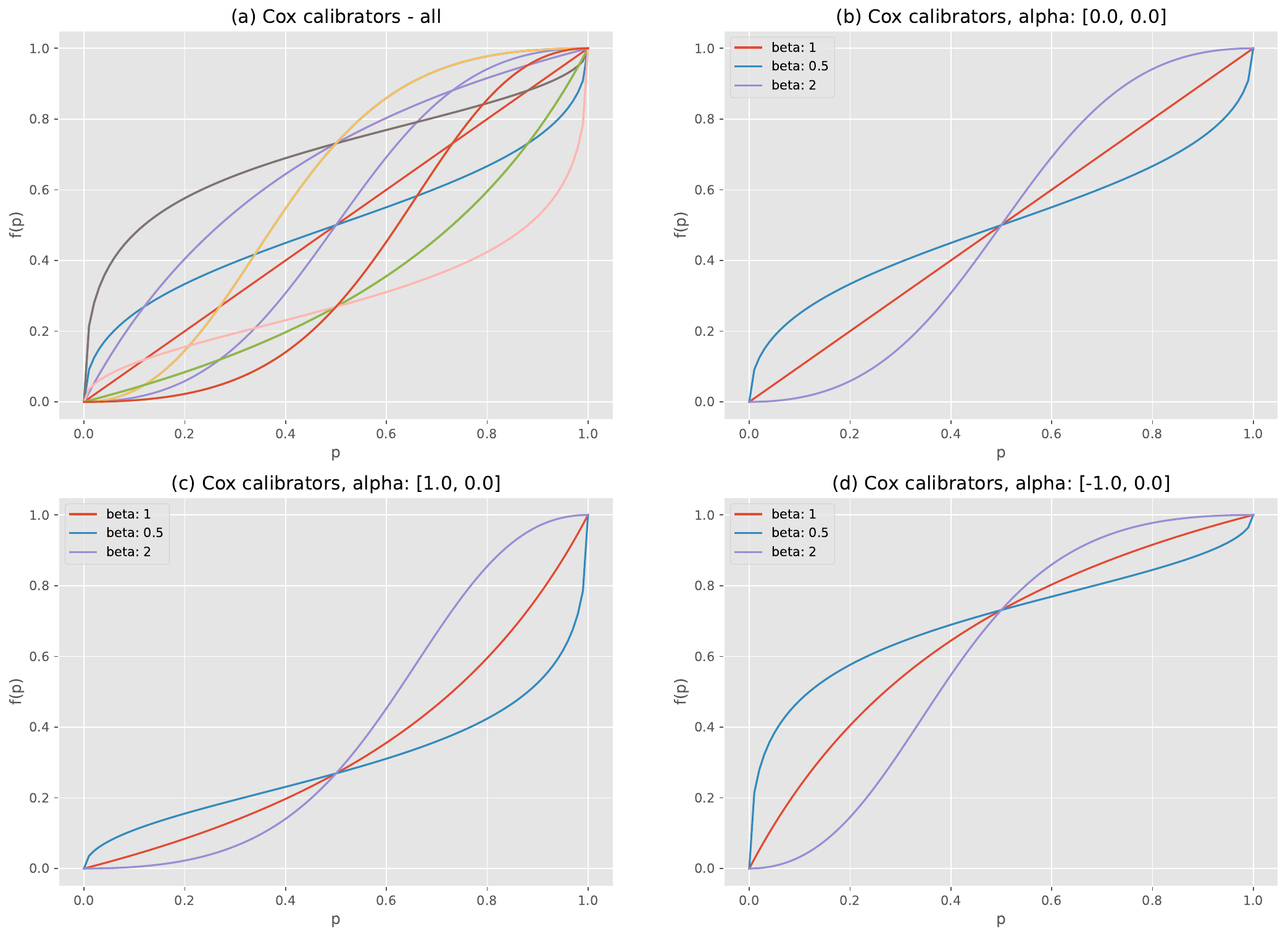}
	\caption{Cox calibrator functions for binary class datasets}
	\label{fig:figure_1}
\end{figure}

Figure \ref{fig:figure_1} illustrates Cox calibrating functions for the binary class problem with subplot (b) showing  the case of $\alpha(y) = [0.0, 0.0]$, subplot (c) showing the case of $\alpha(y) = [1.0, 0.0]$ and subplot (d) showing the case of $\alpha(y) = [-1.0, 0.0]$ separately for clarity. By setting $\alpha$ to zero for all $y$ (subplot (a)), we obtain the one parameter subfamily where $\beta = 1$ is the neutral value indicating no calibration, $\beta = 0.5$ corrects the overconfidence of underlying base predictive system and $\beta=2.0$ the under-confidence.  The equivalent plots in  \ref{fig:figure_1}  (c) and (d) show the Cox calibrator plots aimed for correcting any possible asymmetric over and under-confidence of the base predictive system.

In contrast to a number of post-hoc calibration algorithms, which are applied to the base predictive system only once post initial training, our method of calibration is applied online.  Since we do not know in advance which of the calibrating functions will be best for a given base predictive system, we use a betting framework to make this decision in an online fashion. To do so, we define a \emph{probability process} as a function mapping any finite sequence of observations to the product $B_{p_1}(y_1)\dots  B_{p_n}(y_n)$ (i.e., the probability attached to this sequence by the predictive system), where $n$ is the number of observations in the sequence, $y_1,\dots y_n$ are their labels,  $p_1,\dots p_n$ are the predictions for those observations and $B_p$,
is the Bernoulli distribution on ${0,1}$ with $B_p({1}) = p$.  We regard this probability process as the capital process of a player playing a betting game where their capital cannot go up, and for it not to go down they have to predict with the probability measure concentrated on the true outcome. Next we define a test (always positive) martingale with respect to the base predictive system as the ratio of a probability process of a given calibrating function  to the probability process of the base predictive system and vice versa. The relative size of the test martingale  determines the relative weight we place on the individual calibrating functions applied to the base predictive system. 

%\begin{algorithm}
%  \caption{Composite Jumper predictor {\\
		%	\null\qquad}{}($(p_1,p_2,\dots)\mapsto(p'_1,p'_2,\dots)$)} 
%	\label{alg:CJ-predictor}
%\begin{algorithmic}[1]
%	\State $P:=\pi$\label{ln:prior-1}
%	\State $A^J_{\theta}:=\frac{1-\pi}{\left|\mathbf{J}\right|}1_{\theta=0}$ for all $J\in\mathbf{J}$ and $\theta\in\Theta$\label{ln:prior-2}
%	\For{$n=1,2,\dots$:}
%	\For{$J\in\mathbf{J}$:}
%	\State $A:=\sum_{\theta}A^J_{\theta}$
%	\For{$\theta\in\Theta$:}
%	$A^J_{\theta} := (1-J)A^J_{\theta} + A J/\left|\Theta\right|$\label{ln:recursion-1}
%	\EndFor
%	\EndFor
%	\State $p'_n := p_n P + \sum_{J,\theta} f_{\theta}(p_n) A^J_{\theta}$\label{ln:result}
%	\State $P := P B_{p_n}(\{y_n\})$\label{ln:recursion-2}
%	\For{$J\in\mathbf{J}$ and $\theta\in\Theta$:}
%	$A^J_{\theta} := A^J_{\theta} B_{f_{\theta}(p_n)}(\{y_n\})$\label{ln:recursion-3}
%	\EndFor
%	\State $C := P + \sum_{J,\theta} A^J_{\theta}$\label{ln:C}
%	\State $P := P/C$\label{ln:recursion-4}
%	\For{$J\in\mathbf{J}$ and $\theta\in\Theta$:}
%	$A^J_{\theta} := A^J_{\theta} / C$\label{ln:recursion-5}
%	\EndFor
%	\EndFor
%	\end{algorithmic}
%\end{algorithm}

\begin{algorithm}[h]
\label{alg:CJ-predictor}
\caption{Composite Jumper predictor \(( (p_1, p_2, \ldots) \mapsto (p'_1, p'_2, \ldots) )\)}
\begin{algorithmic}[1]
	
	\STATE $P := \pi$
	\STATE $A^J_\theta := \frac{1 - \pi}{|\mathbf{J}|}  \mathbb{1}_{\theta = 0}$ \hfill \COMMENT{for all $J \in \mathbf{J}$ and $\theta \in \Theta$}
	
	\FOR{$n = 1, 2, \ldots$}
	\FORALL{$J \in \mathbf{J}$}
	\STATE $A := \sum_{\theta} A^J_\theta$
	\FORALL{$\theta \in \Theta$}
	\STATE $A^J_\theta := (1 - J)A^J_\theta + AJ / |\Theta|$
	\ENDFOR
	\ENDFOR
	\STATE $p'_n := p_n P + \sum_{J, \theta} f_\theta(p_n) A^J_\theta$
	\STATE $P := P \cdot B_{p_n}(\{ y_n \})$
	\FORALL{$J \in \mathbf{J}$ and $\theta \in \Theta$}
	\STATE $A^J_\theta := A^J_\theta \cdot B_{f_\theta(p_n)}(\{ y_n \})$
	\ENDFOR
	\STATE $C := P + \sum_{J, \theta} A^J_\theta$
	\STATE $P := P / C$
	\FORALL{$J \in \mathbf{J}$ and $\theta \in \Theta$}
	\STATE $A^J_\theta := A^J_\theta / C$
	\ENDFOR
	\ENDFOR
	
\end{algorithmic}
\end{algorithm}

The intuition behind method above is that we start from a base predictive system, design a way of gambling against it (a test martingale) and then use the martingale (applying the idea of “tracking the best expert” (\cite{herbster1998tracking}) as protection against the kind of changes that the test martingale benefits from. If and when those changes happen, the product of base probability process and the test martingale (defined as the protected probability process) outperforms the base probability process in terms of the log loss function.  The method further makes use of the concept of  \emph{Composite Jumper} martingale with a two-parameter set $\pi \in (0,1)$ and a finite set $\mathbf{J}$ of non-zero jumping rates $J$.  Each jumping rate characterises the transition function  between maintaining the same state with probability $1 - J$ and, with probability $J$, choosing a new state from the uniform probability measure on $\mathbf{\Theta}$. The full method, summarised in Algorithm 1 and described fully in \cite{vovk2021protected} is refereed to as \emph{protected probabilistic classification}. By continually correcting the base predictive system using a calibration function which obtains the highest capital under a martingale betting process described in Algorithm 1 we calibrate (protect) in an adaptive fashion. The cost of such a calibration is given by $\log \frac{1}{\pi}$ (\citet[Section~4]{vovk2021protected}).

\section {Python \texttt{protected-classification} package}
\sloppy
In order to enable the method of protected probabilistic classification more widely available we introduce an open source package \texttt{protected-classification} shared at \texttt{PyPi}\footnote{\url{https://pypi.org/project/protected-classification/}}  under the MIT License. The package is flexible and caters for binary and multi-class settings as well as settings compatible with streaming data problems analogous to those used by a popular machine learning package \texttt{river}~\cite{montiel2021river}. \par
Further details and  instruction on  use of the package are available in the \texttt{github}\footnote{\url{https://github.com/ip200/protected-classification}} repository. Here we illustrate the key features by applying it to the \texttt{Bank marketing} dataset \cite{moro2014data} from the UCI Machine Learning Repository, thereby ensuring consistency with previous reported work in \cite{vovk2021protected}.
The dataset represents results of a marketing campaign of a Portuguese banking institution and the goal is to predict whether a client will subscribe to a term deposit, making it useful for classification tasks.  We take the first 10,000 observations as the training set and use the remaining 35,211 as the test set, without shuffling the data. We train the \texttt{scikit-learn} random forest algorithm with a random seed of \texttt{2021} and all hyper-parameters set to default except for \texttt{no\_estimators} which was set to a value of 1000. This is in slight contrast to the equivalent experiment in  \cite{vovk2021protected}, but it does avoid the need for truncation of underlying random forest probabilities as described in Section 5, Equation (14). We use the same set of Cox calibrating function parameters as described in Section 3, namely $\beta \in \{1, 0.5, 2\}$ and $\alpha(y) \in \{0,  1,  -1\}$ for each class label $y$ and  set the Composite Jumper jumping rates to $\textbf{J} = \{10^{-2}, 10^{-3}, 10^{-4}\}$. These parameters are set as default in the \texttt{protected-classification} library. As this is a binary classification problem the full set of Cox calibrating functions are the same as those depicted in Figure \ref{fig:figure_1}.
\par
\begin{figure}[h]
\centering
\includegraphics[width=\textwidth]{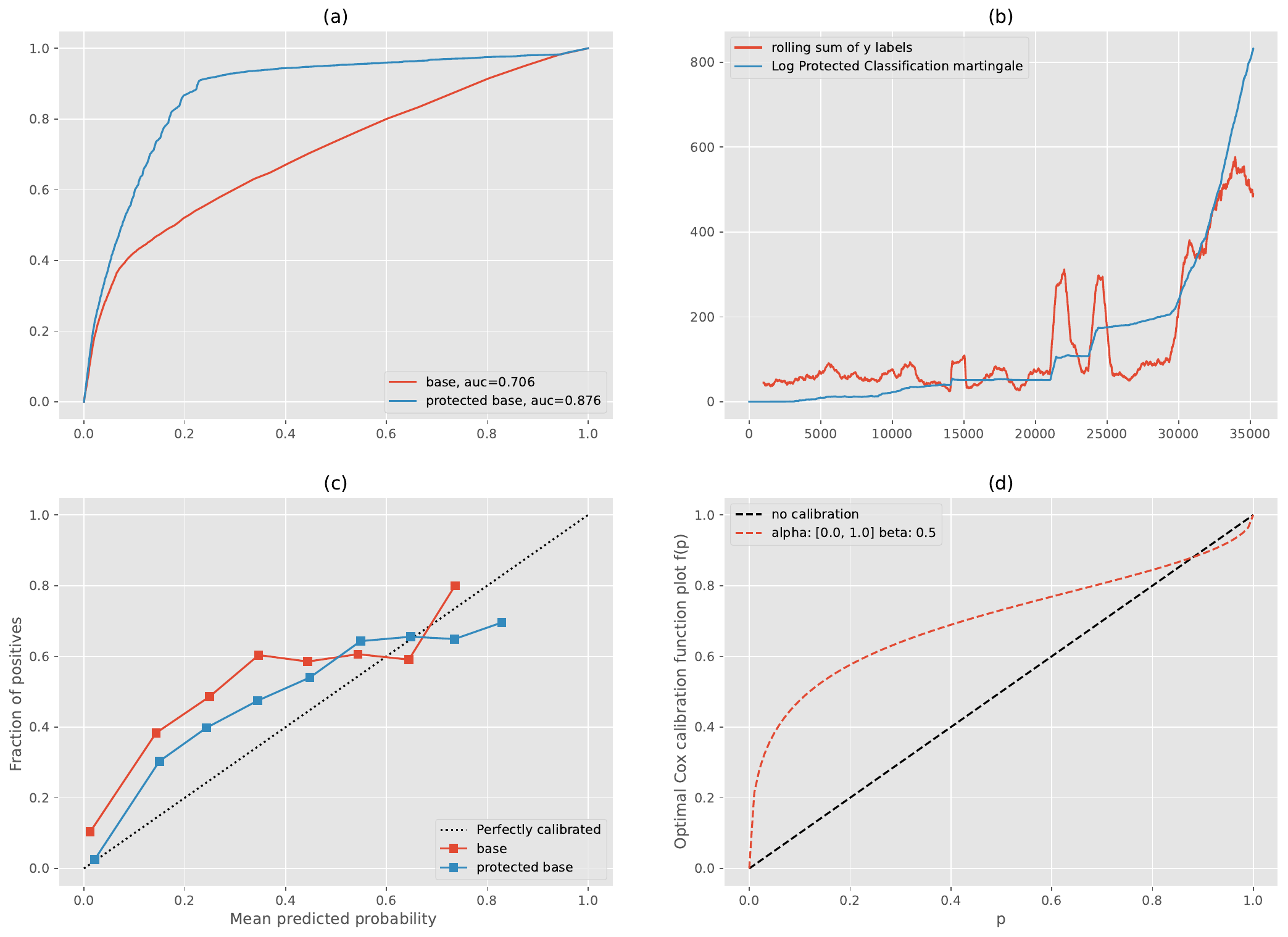}
\caption{(a) ROC/AUC}
\label{fig:figure_2}
\end{figure}

 Figure \ref{fig:figure_2} shows the results of application of the  \texttt{protected-classification} algorithm (\emph{protected base}) to the test set results generated by the underlying random forest algorithm (\emph{base}).  Figure \ref{fig:figure_2}(a) shows comparison of the ROC/AUC curves between the two with a clear outperformance of \emph{protected base} over \emph{base}. Figure \ref{fig:figure_2}(b) shows the relative increase of the positive class label in the later part of the test set and the resulting increase of the test martingale indicating the outperformance of one or more of the Cox calibrator functions vs. the base algorithm in terms of log loss. A closer inspection of the individual Cox calibrator function with the highest final martingale value is shown in Figure \ref{fig:figure_2}(d)  - its shape is similar to the shape of the underlying random forest algorithm calibration plot shown in Figure \ref{fig:figure_2}(c), indicating that the algorithm correctly tries to protect against the miscalibration caused in part by the distribution shift of the labels shown in Figure \ref{fig:figure_2}(b). 

\begin{center}
\begin{table}
	\centering
	\small
	\label{tbl:bankmarketing}
	\caption{Performance comparison between for the underlying RF algorithm and its protected version on the \texttt{Bank Marketing} dataset}
	\vspace{0.5cm}
	\begin{tabular}{l|rrr}
		\toprule
		& accuracy & Brier loss & log loss  \\
		\midrule
		base & 0.861 & 	0.118 &  0.589 \\
		protected base & \textbf{0.871} & \textbf{0.091} & \textbf{0.363}  \\
		\midrule
		\bottomrule
	\end{tabular}
\end{table}
\end{center}

\newcommand*\wrapletters[1]{\wr@pletters#1\@nil}
\def\wr@pletters#1#2\@nil{#1\allowbreak\if&#2&\else\wr@pletters#2\@nil\fi}

Table 1 shows the relative outperformance of the protected calibration over the base algorithm across accuracy, Brier loss and log loss (with \textbf{bold} indicating a better outcome, the convention used in the remainder of this chaoter). The code illustrating results in this section can be found in the  \texttt{github} repository under \texttt{notebooks/protected-classification}. The same repository shows examples of protected classification applied in three other settings - \emph{batch} mode~(\texttt{notebooks/protected-batch}), where the calibration is applied to a validation set only, analogous to standard post-hoc calibration algorithms such as Platt scaling (\cite{platt1999probabilistic}) or isotonic regression (\cite{zadrozny2001learning}), \emph{multiclass}~(\texttt{\wrapletters{notebooks/protected-multiclass}}), where we illustrate the application of the algorithm to multiclass classification problems and \emph{streaming}~(\texttt{notebooks/protected-streaming}), where we illustrate the application of the algorithm to streaming data problems compatible with \texttt{river} library~\cite{montiel2021river}. The latter two applications will be further illustrated in the next section.

\section{Experimental results}

The Experimental results section consists of three sets of experiments. In the first section we compare the performance of \texttt{protected-classification} algorithm vs. traditional machine learning algorithms and existing post-hoc calibration methods in binary and multi-class settings under dataset shift, i.e., where the data in the test set is induced to have a different distribution to that in the training/calibration sets.  In the second section we compare our method with a recently reported equivalent technique of \emph{Online Platt Scaling with Calibeating}~\cite{gupta2023online}. In the final section, we apply our algorithm within streaming data problems, comparing its effectiveness relative to other state-of-the art techniques in this setting. 

\subsection{Experiments under dataset shift}

\subsubsection{Artificial datasets}
In the set of experiments using synthetically generated data,  we used the \texttt{scikit-learn} functionality to create a set of datasets. Each dataset consists of 20 input features, and contains 2000 training/calibration examples and 1000 test examples. We consider 2, 3, 5 and 10-class problems, in each case generating five independent datasets using a different random seed.

In order to carry our protected-classification experiments under dataset shift, we consider four scenarios: the \emph{unperturbed} dataset, where we assume that the training and test have the same probability distribution, the \emph{concept shift}, where we swap the binary class labels for the second half of the test dataset only (last 500 examples), the \emph{X-imbalance}, where we simulate the change in the underlying feature distribution by including examples only where one of the relevant numerical features has a value of less than 0 for the last 500 examples, and finally the \emph{y-imbalance}, where we simulate the change in the frequency of labels and exclude the examples whose labels are equal to 0 for the last 500 test points. 

We use 5 different underlying classifiers: Gaussian na\"ive Bayes, logistic regression, random forests,  support vector machines and gradient boosted trees, all sourced from \texttt{scikit-learn} using the default set of hyperparameters. We consider each underlying algorithm in its base form $(b)$ and also calibrated using \emph{Platt scaling} $(s)$, \emph{isotonic Regression} $(i)$ and \emph{Venn-ABERS} calibration \cite{vovk2015large} $(v)$ using two-fold cross-validation for all. The former two calibration methods were carried out using the \texttt{scikit-learn calibration} functionality and the latter using the \texttt{venn-abers} package from PyPi~\footnote{\url{pypi.org/project/venn-abers}}. Our goal is to asses whether the protected classification algorithm can outperform the base and calibrated underlying algorithms on the test set under the log loss, Brier loss and expected calibration error (ECE, defined as in ~\cite{kumar2019calibration}) when the test examples are exchangable (where the modification to the test set is applied only to the last 500 examples and the order of the test set examples is then randomly permuted).  The code to reproduce all experiments can be found in the Online Appendix\footnote{\url{https://github.com/ip200/protected-classification-experiments}}. Tables 2-5 show the average ECE for each classifier, number of classes and base or calibration algorithm with and without protected classification applied on top of the underlying method.

\begin{table}
\caption{Expected calibration error (ECE) - \emph{unperturbed} dataset experiments}
\tiny
\centering
\vspace{0.5cm}
\begin{tabular}{l|l|rrrr|rrrr}
	\toprule
	&  & \multicolumn{4}{c}{standard} & \multicolumn{4}{c}{protected} \\
	&  & b & i & s & v & b & i & s & v \\
	classes & classifier &  &  &  &  &  &  &  &  \\
	\midrule
	\midrule
	\multirow[c]{5}{*}{2} & Logistic & 0.043 & \textbf{0.000} & 0.047 & 0.041 & 0.042 & 0.000 & 0.046 & 0.000 \\
	& Naive Bayes & 0.108 & 0.076 & 0.083 & \textbf{0.000} & 0.046 & 0.000 & 0.075 & 0.000 \\
	& Random forest & 0.056 & 0.033 & \textbf{0.000} & 0.038 & 0.036 & 0.008 & 0.000 & 0.000 \\
	& SVM & 0.050 & 0.053 & 0.050 & \textbf{0.000} & 0.050 & 0.000 & 0.050 & 0.000 \\
	& XGBoost & 0.017 & 0.031 & 0.029 & \textbf{0.000} & 0.023 & 0.000 & 0.030 & 0.000 \\
	\midrule
	\multirow[c]{5}{*}{3} & Logistic & 0.027 & 0.034 & 0.055 & \textbf{0.021} & 0.026 & 0.035 & 0.056 & 0.025 \\
	& Naive Bayes & 0.118 & \textbf{0.026} & 0.092 & 0.027 & 0.050 & 0.030 & 0.090 & 0.029 \\
	& Random forest & 0.055 & 0.007 & 0.034 & \textbf{0.000} & 0.035 & 0.006 & 0.032 & 0.000 \\
	& SVM & 0.034 & \textbf{0.022} & 0.034 & 0.023 & 0.035 & 0.024 & 0.035 & 0.023 \\
	& XGBoost & 0.023 & 0.024 & 0.028 & 0.004 & 0.019 & 0.025 & 0.029 & \textbf{0.000} \\
	\midrule
	\multirow[c]{5}{*}{5} & Logistic & 0.042 & 0.063 & 0.068 & 0.040 & 0.040 & 0.047 & 0.047 & \textbf{0.039} \\
	& Naive Bayes & 0.114 & 0.048 & 0.053 & 0.044 & 0.056 & 0.048 & 0.051 & \textbf{0.044} \\
	& Random forest & 0.080 & 0.041 & 0.041 & 0.033 & \textbf{0.017} & 0.040 & 0.041 & 0.031 \\
	& SVM & 0.066 & 0.039 & 0.066 & 0.038 & 0.043 & 0.039 & 0.043 & \textbf{0.038} \\
	& XGBoost & 0.030 & 0.080 & 0.085 & 0.024 & 0.028 & 0.025 & 0.028 & \textbf{0.024} \\
	\midrule
	\multirow[c]{5}{*}{10} & Logistic & 0.037 & \textbf{0.029} & 0.037 & 0.031 & 0.036 & 0.029 & 0.036 & 0.031 \\
	& Naive Bayes & 0.098 & \textbf{0.036} & 0.036 & 0.037 & 0.042 & 0.036 & 0.037 & 0.038 \\
	& Random forest & 0.078 & 0.043 & 0.039 & 0.038 & \textbf{0.018} & 0.043 & 0.036 & 0.037 \\
	& SVM & 0.038 & 0.021 & 0.038 & 0.031 & 0.038 & \textbf{0.021} & 0.038 & 0.031 \\
	& XGBoost & 0.019 & 0.054 & 0.065 & 0.034 & 0.019 & \textbf{0.018} & 0.020 & 0.035 \\\midrule\ & Average & 0.057 & 0.038 & 0.049 & 0.025 & 0.035 & 0.024 & 0.041 & \textbf{0.021} \\
	\bottomrule
\end{tabular}
\end{table}

\begin{table}
\caption{Expected calibration error (ECE) - \emph{concept shift} dataset experiments}
\tiny
\centering
\vspace{0.5cm}
\begin{tabular}{l|l|rrrr|rrrr}
	\toprule
	&  & \multicolumn{4}{c}{standard} & \multicolumn{4}{c}{protected} \\
	&  & b & i & s & v & b & i & s & v \\
	classes & classifier &  &  &  &  &  &  &  &  \\
	\midrule
	\midrule
	\multirow[c]{5}{*}{2} & Logistic & 0.373 & 0.384 & 0.367 & 0.371 & 0.278 & 0.278 & 0.270 & \textbf{0.267} \\
	& Naive Bayes & 0.462 & 0.388 & 0.393 & 0.392 & 0.383 & 0.301 & \textbf{0.255} & 0.288 \\
	& Random forest & 0.429 & 0.456 & 0.446 & 0.443 & 0.333 & 0.357 & \textbf{0.323} & 0.337 \\
	& SVM & 0.366 & 0.382 & 0.366 & 0.381 & \textbf{0.266} & 0.275 & 0.266 & 0.266 \\
	& XGBoost & 0.456 & 0.451 & 0.454 & 0.447 & 0.338 & 0.375 & \textbf{0.330} & 0.344 \\
	\midrule
	\multirow[c]{5}{*}{3} & Logistic & 0.210 & 0.187 & 0.168 & 0.198 & 0.138 & 0.121 & \textbf{0.115} & 0.126 \\
	& Naive Bayes & 0.258 & 0.188 & 0.202 & 0.171 & 0.195 & 0.126 & 0.114 & \textbf{0.113} \\
	& Random forest & 0.201 & 0.228 & 0.217 & 0.225 & 0.143 & 0.144 & \textbf{0.134} & 0.144 \\
	& SVM & 0.176 & 0.214 & 0.176 & 0.203 & \textbf{0.116} & 0.129 & 0.116 & 0.127 \\
	& XGBoost & 0.232 & 0.217 & 0.212 & 0.220 & 0.145 & 0.143 & \textbf{0.138} & 0.139 \\
	\midrule
	\multirow[c]{5}{*}{5} & Logistic & 0.068 & 0.058 & \textbf{0.055} & 0.074 & 0.066 & 0.055 & 0.056 & 0.067 \\
	& Naive Bayes & 0.156 & 0.086 & 0.084 & 0.096 & 0.086 & \textbf{0.074} & 0.080 & 0.088 \\
	& Random forest & 0.113 & 0.114 & 0.103 & 0.115 & 0.105 & 0.104 & \textbf{0.090} & 0.099 \\
	& SVM & \textbf{0.054} & 0.059 & 0.054 & 0.073 & 0.056 & 0.061 & 0.056 & 0.068 \\
	& XGBoost & 0.121 & 0.093 & 0.093 & 0.110 & 0.101 & 0.094 & \textbf{0.089} & 0.093 \\
	\midrule
	\multirow[c]{5}{*}{10} & Logistic & 0.048 & \textbf{0.034} & 0.038 & 0.045 & 0.042 & 0.034 & 0.036 & 0.044 \\
	& Naive Bayes & 0.111 & 0.052 & \textbf{0.042} & 0.048 & 0.056 & 0.049 & 0.042 & 0.047 \\
	& Random forest & 0.066 & 0.048 & 0.046 & 0.043 & \textbf{0.038} & 0.045 & 0.043 & 0.043 \\
	& SVM & 0.041 & \textbf{0.032} & 0.041 & 0.045 & 0.041 & 0.032 & 0.041 & 0.044 \\
	& XGBoost & 0.041 & 0.058 & 0.060 & 0.043 & 0.041 & 0.042 & \textbf{0.037} & 0.040 \\\midrule\ & Average & 0.199 & 0.187 & 0.181 & 0.187 & 0.148 & 0.142 & \textbf{0.132} & 0.139 \\
	\bottomrule
\end{tabular}
\end{table}

\begin{table}
\caption{Expected calibration error (ECE) - \emph{X-imbalance} dataset experiments}
\tiny
\centering
\vspace{0.5cm}
\begin{tabular}{l|l|rrrr|rrrr}
	\toprule
	&  & \multicolumn{4}{c}{standard} & \multicolumn{4}{c}{protected} \\
	&  & b & i & s & v & b & i & s & v \\
	classes & classifier &  &  &  &  &  &  &  &  \\
	\midrule
	\midrule
	\multirow[c]{5}{*}{2} & Logistic & 0.045 & \textbf{0.000} & 0.044 & 0.036 & 0.044 & 0.000 & 0.044 & 0.000 \\
	& Naive Bayes & 0.110 & 0.014 & 0.084 & \textbf{0.000} & 0.026 & 0.000 & 0.075 & 0.000 \\
	& Random forest & 0.057 & \textbf{0.000} & 0.022 & 0.000 & 0.041 & 0.000 & 0.016 & 0.015 \\
	& SVM & 0.049 & 0.047 & 0.049 & 0.047 & 0.051 & \textbf{0.000} & 0.051 & 0.000 \\
	& XGBoost & 0.023 & \textbf{0.000} & 0.011 & 0.000 & 0.027 & 0.000 & 0.008 & 0.000 \\
	\midrule
	\multirow[c]{5}{*}{3} & Logistic & 0.167 & 0.154 & 0.134 & 0.160 & 0.116 & \textbf{0.104} & 0.107 & 0.112 \\
	& Naive Bayes & 0.213 & 0.123 & 0.179 & 0.115 & 0.121 & \textbf{0.081} & 0.099 & 0.108 \\
	& Random forest & 0.183 & 0.179 & 0.213 & 0.140 & 0.149 & \textbf{0.132} & 0.144 & 0.133 \\
	& SVM & 0.130 & 0.171 & 0.130 & 0.164 & \textbf{0.108} & 0.116 & 0.108 & 0.115 \\
	& XGBoost & 0.208 & 0.195 & 0.197 & 0.215 & \textbf{0.117} & 0.124 & 0.139 & 0.143 \\
	\midrule
	\multirow[c]{5}{*}{5} & Logistic & 0.036 & 0.064 & 0.067 & 0.036 & 0.035 & 0.051 & 0.040 & \textbf{0.034} \\
	& Naive Bayes & 0.113 & 0.053 & 0.052 & 0.042 & 0.049 & 0.053 & 0.051 & \textbf{0.041} \\
	& Random forest & 0.079 & 0.041 & 0.045 & 0.031 & \textbf{0.016} & 0.039 & 0.042 & 0.026 \\
	& SVM & 0.064 & 0.042 & 0.064 & 0.032 & 0.056 & 0.040 & 0.056 & \textbf{0.032} \\
	& XGBoost & 0.030 & 0.081 & 0.088 & 0.027 & 0.029 & \textbf{0.018} & 0.020 & 0.021 \\
	\midrule
	\multirow[c]{5}{*}{10} & Logistic & 0.031 & 0.022 & 0.030 & 0.028 & 0.030 & \textbf{0.021} & 0.029 & 0.028 \\
	& Naive Bayes & 0.098 & \textbf{0.028} & 0.034 & 0.032 & 0.041 & 0.029 & 0.034 & 0.032 \\
	& Random forest & 0.071 & 0.037 & 0.040 & 0.030 & \textbf{0.014} & 0.033 & 0.036 & 0.030 \\
	& SVM & 0.035 & \textbf{0.022} & 0.035 & 0.033 & 0.035 & 0.022 & 0.035 & 0.032 \\
	& XGBoost & 0.020 & 0.051 & 0.059 & 0.031 & 0.020 & 0.022 & \textbf{0.019} & 0.031 \\\midrule\ & Average & 0.088 & 0.066 & 0.079 & 0.060 & 0.056 & \textbf{0.044} & 0.058 & 0.047 \\
	\bottomrule
\end{tabular}
\end{table}

\begin{table}
\caption{Expected calibration error (ECE) - \emph{y-imbalance} dataset experiments}
\tiny
\centering
\vspace{0.5cm}
\begin{tabular}{l|l|rrrr|rrrr}
	\toprule
	&  & \multicolumn{4}{c}{standard} & \multicolumn{4}{c}{protected} \\
	&  & b & i & s & v & b & i & s & v \\
	classes & classifier &  &  &  &  &  &  &  &  \\
	\midrule
	\midrule
	\multirow[c]{5}{*}{2} & Logistic & 0.088 & 0.068 & 0.092 & 0.074 & 0.026 & \textbf{0.010} & 0.028 & 0.011 \\
	& Naive Bayes & 0.105 & 0.110 & 0.114 & 0.076 & 0.059 & 0.035 & 0.040 & \textbf{0.032} \\
	& Random forest & 0.019 & \textbf{0.000} & 0.034 & 0.027 & 0.029 & 0.022 & 0.025 & 0.024 \\
	& SVM & 0.090 & 0.085 & 0.090 & 0.069 & 0.034 & \textbf{0.009} & 0.034 & 0.031 \\
	& XGBoost & 0.037 & 0.040 & 0.048 & 0.053 & 0.030 & \textbf{0.018} & 0.040 & 0.040 \\
	\midrule
	\multirow[c]{5}{*}{3} & Logistic & 0.089 & 0.094 & 0.106 & 0.085 & 0.058 & 0.069 & 0.058 & \textbf{0.047} \\
	& Naive Bayes & 0.149 & 0.092 & 0.124 & 0.092 & 0.060 & 0.047 & 0.065 & \textbf{0.040} \\
	& Random forest & 0.054 & 0.074 & 0.072 & 0.088 & 0.051 & \textbf{0.043} & 0.052 & 0.057 \\
	& SVM & 0.089 & 0.074 & 0.089 & 0.081 & 0.074 & 0.053 & 0.074 & \textbf{0.044} \\
	& XGBoost & 0.065 & 0.053 & 0.054 & 0.077 & 0.040 & \textbf{0.038} & 0.053 & 0.050 \\
	\midrule
	\multirow[c]{5}{*}{5} & Logistic & 0.077 & 0.095 & 0.101 & 0.069 & 0.055 & 0.048 & \textbf{0.048} & 0.060 \\
	& Naive Bayes & 0.112 & 0.076 & 0.082 & 0.068 & \textbf{0.059} & 0.067 & 0.061 & 0.060 \\
	& Random forest & 0.067 & 0.041 & 0.049 & 0.038 & \textbf{0.033} & 0.037 & 0.045 & 0.037 \\
	& SVM & 0.095 & 0.072 & 0.095 & 0.067 & \textbf{0.049} & 0.053 & 0.049 & 0.056 \\
	& XGBoost & 0.036 & 0.083 & 0.097 & 0.039 & 0.036 & 0.026 & \textbf{0.021} & 0.033 \\
	\midrule
	\multirow[c]{5}{*}{10} & Logistic & 0.046 & \textbf{0.043} & 0.051 & 0.045 & 0.045 & 0.043 & 0.051 & 0.045 \\
	& Naive Bayes & 0.104 & 0.047 & 0.050 & 0.046 & 0.050 & 0.046 & 0.050 & \textbf{0.045} \\
	& Random forest & 0.074 & 0.049 & 0.049 & 0.038 & \textbf{0.024} & 0.039 & 0.043 & 0.040 \\
	& SVM & 0.052 & 0.045 & 0.052 & 0.042 & 0.050 & 0.043 & 0.050 & \textbf{0.042} \\
	& XGBoost & 0.028 & 0.064 & 0.074 & 0.047 & \textbf{0.028} & 0.030 & 0.028 & 0.046 \\\midrule\ & Average & 0.074 & 0.065 & 0.076 & 0.061 & 0.045 & \textbf{0.039} & 0.046 & 0.042 \\
	\bottomrule
\end{tabular}
\end{table}

The protected classification results for the $unperturbed$ dataset are  mostly in line with the base algorithm results, with the mild improvement largely a rebuilt of the better calibration of the underlying algorithms themselves rather than protection against any dataset shift. In other experiment sets with dataset shift, the protected calibration algorithm generally results in a lower calibration error across most perturbations with protected classification applied on top of one of the calibration methods yielding the best results on average. The results are more significant for lower number of classes. It is important to note that the ECE measure used in our experiments sourced from the \texttt{uncertainty-calibration} package \cite{kumar2019calibration} measures the calibration error of all classes (per-class calibration error), rather than just the equivalent for the top-prediction. For the multi-class experiments the former measure is a stronger notion of calibration. 

\subsubsection{Distorted image classification}

In this set of experiments we use the standard MNIST digit classification dataset which we analyse using a standard neural network (3-layer CNN with pooling), commonly used as an architecture for MNIST digit classification.   The standard MNIST dataset consists of 50,000 training points and 10,000 test points. The idea of these protected calibration experiments is to apply incremental distortion transformations to the objects in the test set in order to shift the distribution of data used by the CNN classifier to make a prediction. The transformations made are image rotation (from 10 degrees to a maximum of 90 degrees), after the paper \emph{Failing Loudly: An Empirical Study of Methods for Detecting Dataset Shift} \cite{rabanser2019failing}. Only the last 5000 examples are distorted, each of the experiments for each transformation was repeated 5 times and we report the average results. We show the performance of the underlying neural network under ECE, Brier and log loss without (labelled $b$) and with a commonly used calibration method for neural networks, \emph{temperature scaling} after \citet{guo2017calibration} (labelled $t$).

\begin{table}[H]
\caption{MNIST distorted image dataset - Brier loss}
\centering
\vspace{0.5cm}
\small
\begin{tabular}{l|rr|rr||rr|rr}
	\toprule
	& \multicolumn{2}{c}{standard} & \multicolumn{2}{c}{protected} \\
	& b & t & b & t \\
	\midrule
	10 rotation & 0.0712 & 0.0739 & \textbf{0.0677} & 0.0697 \\
	30 rotation & 0.1261 & 0.1335 & \textbf{0.1225} & 0.1227 \\
	50 rotation & 0.2656 & 0.2841 & 0.2535 & \textbf{0.2458} \\
	90 rotation & 0.4486 & 0.4796 & \textbf{0.3916} & 0.4048 \\
	\midrule
	Average & 0.2279 & 0.2428 & \textbf{0.2088} & 0.2108 \\
	\bottomrule
\end{tabular}
\end{table}

\begin{table}[H]
\caption{MNIST distorted image dataset - log loss}
\centering
\vspace{0.5cm}
\small
\begin{tabular}{l|rr|rr||rr|rr}
	\toprule
	& \multicolumn{2}{c}{standard} & \multicolumn{2}{c}{protected} \\
	& b & t & b & t \\
	\midrule
	10 rotation & 0.1517 & 0.1711 & \textbf{0.1491} & 0.1573 \\
	30 rotation & 0.2773 & 0.3310 & 0.2777 & \textbf{0.2744} \\
	50 rotation & 0.6912 & 0.8327 & 0.6552 & \textbf{0.5791} \\
	90 rotation & 1.5246 & 1.8207 & \textbf{1.0706} & 1.0974 \\
	\midrule
	Average & 0.6612 & 0.7889 & 0.5382 & \textbf{0.5271} \\
	\bottomrule
\end{tabular}
\end{table}

\begin{table}[H]
\caption{MNIST distorted image dataset - ECE}
\centering
\vspace{0.5cm}
\small
\begin{tabular}{l|rr|rr||rr|rr}
	\toprule
	& \multicolumn{2}{c}{standard} & \multicolumn{2}{c}{protected} \\
	& b & t & b & t \\
	\midrule
	10 rotation & 0.0198 & 0.0193 & 0.0144 & \textbf{0.0138} \\
	30 rotation & 0.0240 & 0.0260 & \textbf{0.0175} & 0.0176 \\
	50 rotation & 0.0499 & 0.0581 & 0.0377 & \textbf{0.0332} \\
	90 rotation & 0.0806 & 0.0919 & \textbf{0.0452} & 0.0570 \\
	\midrule
	Average & 0.0435 & 0.0488 & \textbf{0.0287} & 0.0304 \\
	\bottomrule
\end{tabular}
\end{table}

The results suggest that protected classification significantly improves the calibration and log and Brier losses with and without temperature scaling with higher improvement for larger distortions. This could be a helpful technique therefore in real-life imaging classifications where some dataset shift is expected to occur post initial training and calibration.

\subsection{Comparison with an alternative online post-hoc calibration method}

In the work so far above we compared the implementation of the protected classification algorithm solely against post-hoc calibration techniques which are applied once post initial training. In this section we directly compare the protected classification algorithm to an equivalent online post-hoc calibration algorithm recently published in \emph{Online Platt Scaling with Calibeating} \cite{gupta2023online}. This method is theoretically guaranteed to be calibrated for adversarial outcome sequences. In particular we follow a set of experiments described in the paper where the authors introduce synthetic drifts in the data based on covariate values, i.e. an instance of covariate drift. We show results here comparing our methods  - applying protected classification to the underlying base algorithm (\emph{BM - protected}), the equivalent but after first calibrating using the Venn-abers method (\emph{Venn-abers protected}), as well as applying protected classification to the authors' underlying method of Online Platt Scaling with hedging (\emph{HOPS - protected}). The experiments were carried out on four real life data sets reported in the paper (\texttt{bank}, \texttt{credit}, \texttt{churn} and \texttt{fetal}) with synthetically induced covariate shift using the authors publicly available code\footnote{\url{https://github.com/aigen/df-posthoc-calibration}}. The underlying algorithms used were random forests \emph{(rf)}, gradient boosted tress $(xb)$ and logistic regression $(lr)$ to which the authors applied five different post-hoc calibration methods described in \citet[Section~4]{gupta2023online}. 

The results in Tables 9 and 10 compare the calibration error (CE, as defined by the authors in the study) between the protected classification method and other equivalent algorithms. The values in each cell represent the average CE for a given algorithm and dataset  across the full length of the sample containing the induced covariate shift.  The protected classification methods are comparable to the methods reported in \cite{gupta2023online}, with protected classification applied on top of OPS + hedging (\emph{HOPS-protected}) yielding the lowest average CE across all datasets and classifiers and the \emph{Venn-abers protected} method the second lowest. The results suggest therefore that applying protected classification in combination with other similar online calibrating algorithms may help to improve calibration further in problems when the underlying data distribution changes in the test set.  

\begin{table}
\caption{CE for the bank and credit datasets, with competing calibration algorithms applied to the underlying classifiers}
\vspace{0.5cm}
\small
\centering
\begin{tabular}{l|rrr|rrr|r|}
	\toprule
	dataset & \multicolumn{3}{c}{bank} & \multicolumn{3}{c}{credit} & average \\
	\midrule
	classifier & lr & rf & xb & lr & rf & xb &  \\
	calibrator &  &  &  &  &  &  &  \\
	\midrule
	Base model (BM) & 0.065 & 0.108 & 0.091 & 0.071 & 0.040 & 0.034 & 0.068 \\
	Fixed-batch Platt scaling (FPS) & 0.025 & 0.051 & 0.043 & 0.037 & 0.039 & 0.024 & 0.036 \\
	Online Platt scaling (OPS) & 0.027 & 0.023 & 0.018 & 0.018 & 0.031 & 0.013 & 0.022 \\
	OPS + tracking (TOPS)& 0.021 & 0.016 & 0.019 & 0.005 & 0.006 & 0.006 & 0.012 \\
	OPS + hedging (HOPS) & 0.022 & 0.017 & 0.015 & 0.012 & 0.021 & 0.021 & 0.018 \\
	Windowed Platt scaling (WPS)& 0.026 & 0.024 & 0.020 & 0.037 & 0.034 & 0.016 & 0.026 \\
	\hdashline
	BM -protected & 0.031 & 0.037 & 0.031 & 0.033 & 0.037 & 0.031 & 0.033 \\
	HOPS - protected & 0.016 & 0.012 & \textbf{0.009} & \textbf{0.005} & \textbf{0.006} & \textbf{0.006} & \textbf{0.009} \\
	Venn-abers protected & \textbf{0.011} & \textbf{0.010} & 0.011 & 0.011 & 0.010 & 0.011 & 0.010 \\
	\bottomrule
\end{tabular}
\end{table}

\begin{table}
\caption{CE for the churn  and fetal datasets, with competing calibration algorithms applied to the underlying classifiers}
\vspace{0.5cm}
\small
\centering
\begin{tabular}{l|rrr|rrr|r|}
	\toprule
	dataset & \multicolumn{3}{c}{churn} & \multicolumn{3}{c}{fetal} & average \\
	\midrule
	classifier & lr & rf & xb & lr & rf & xb &  \\
	calibrator &  &  &  &  &  &  &  \\
	\midrule
	Base model (BM) & 0.050 & 0.039 & 0.055 & 0.139 & 0.160 & 0.129 & 0.095 \\
	Fixed-batch Platt scaling (FPS)  & 0.051 & 0.040 & 0.042 & 0.126 & 0.125 & 0.129 & 0.085 \\
	Online Platt scaling (OPS)& 0.014 & 0.026 & 0.012 & 0.039 & 0.036 & 0.039 & 0.028 \\
	OPS + tracking (TOPS) & 0.013 & 0.015 & 0.010 & 0.038 & 0.022 & 0.029 & 0.021 \\
	OPS + hedging (HOPS) & 0.019 & 0.018 & 0.017 & 0.048 & 0.054 & 0.055 & 0.035 \\
	Windowed Platt scaling (WPS) & 0.042 & 0.033 & 0.034 & 0.084 & 0.071 & 0.081 & 0.057 \\
	\hdashline
	BM -protected  & 0.019 & 0.026 & 0.029 & 0.085 & 0.087 & 0.092 & 0.056 \\
	HOPS - protected & \textbf{0.012} & 0.015 & 0.010 & 0.032 & \textbf{0.019} & \textbf{0.023} & \textbf{0.018} \\
	Venn-abers protected & 0.013 & \textbf{0.012} & \textbf{0.009} & \textbf{0.029} & 0.027 & 0.028 & 0.019 \\
	\bottomrule
\end{tabular}
\end{table}

\subsection{Comparison with online algorithms for streaming data}

In this section we directly compare protected classification applied to algorithms which are designed to learn online. For this we make use of the \texttt{river} library~\cite{montiel2021river}, which provides multiple state-of-the-art learning methods, data generators/transformers, performance metrics and evaluators for different streaming data learning problems. In order to perform equivalent online classification we created a modified version of the protected classification algorithm which is compatible with the \texttt{river} package so that our results are directly comparable.  We illustrate an example of the \texttt{learn\_one} and \texttt{predict\_one} in the \texttt{protected-classification} \texttt{github} repository. 

We apply our modified protected classification method to 8 classification algorithms within the \texttt{river} library for one representative binary dataset (\texttt{Phishing}, a spam message classification problem) and  one multi-class dataset (\texttt{ImageSegments}, an artificial dataset for classifying types of images). Each underlying algorithm is designed to learn online with two in particular (ARF and SRP)  designed to handle datasets with shift. More details on the underlying algorithms and datasets can be found in the online documentation for the \texttt{river} package at \url{https://riverml.xyz/latest/}.

The results in Table 11 show the average log loss  over five different runs with different random seeds for each classifier and dataset combination.

\begin{table}[H]
\caption{Log loss for streaming data experiments}
\vspace{0.5cm}
\centering
\small
\begin{tabular}{l|rr|rr|}
	& \multicolumn{2}{c}{Phishing} & \multicolumn{2}{c}{ImageSegments} \\
	model & base & protected & base & protected \\
	\midrule
	ADWIN & 0.2337 & \textbf{0.1895} & 0.4316 & \textbf{0.4298} \\
	ARF & 0.2336 & \textbf{0.2038} & 0.4598 & \textbf{0.4315} \\
	Bagging & 0.2286 & \textbf{0.1837} & 0.4371 & \textbf{0.4310} \\
	DriftARF & 0.2242 & \textbf{0.2033} & 0.4553 & \textbf{0.4310} \\
	DriftSRP & 0.1778 & \textbf{0.1768} & 0.4394 & \textbf{0.4314} \\
	EFTree & 0.3621 & \textbf{0.3518} & 2.6494 & \textbf{2.2573} \\
	Hoeffding & 0.4333 & \textbf{0.4120} & 0.4302 & \textbf{0.4298} \\
	SRP & 0.1871 & \textbf{0.1858} & 0.4392 & \textbf{0.4312} \\
	\midrule
	Average & 0.2601 & \textbf{0.2383} & 0.7178 & \textbf{0.6591} \\
\end{tabular}
\end{table}

\begin{figure}[hbt!]
\centering
\includegraphics[width=\textwidth]{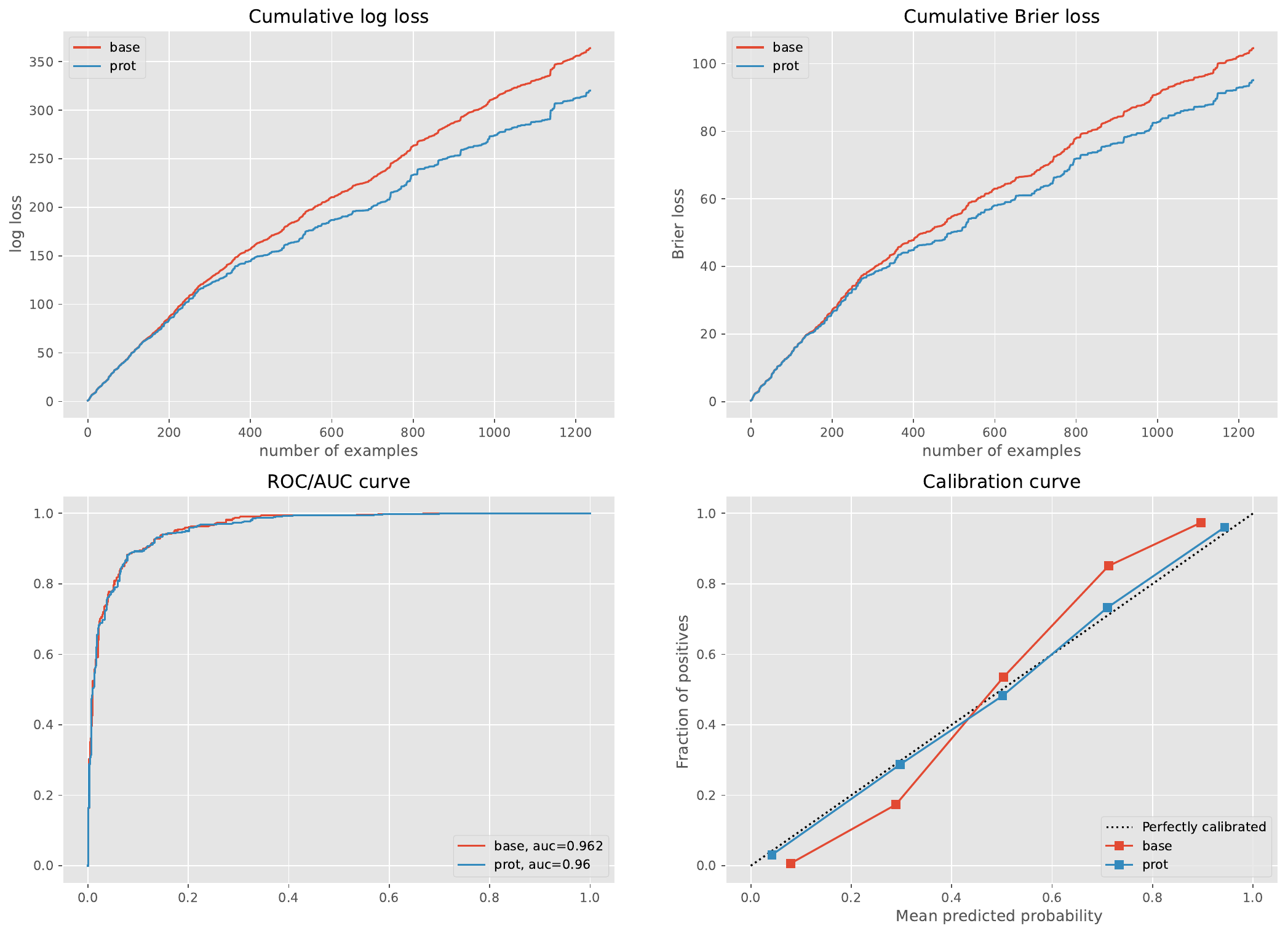}
\caption{Cumulative log loss, Brier loss, total ROC/AUC and calibration curve for the \texttt{ARF} algorithm (base) and the protected classification equivalent (protected) for the \texttt{Phishing} dataset}
\label{fig:figure_3}
\end{figure}

The results suggest that as in the case of comparison in the previous section, applying protected classification to underlying online learning algorithms designed for streaming data can help to produce better calibrated outputs.  Finally, Figure \ref{fig:figure_3} shows an example of the \texttt{protected-classification} applied on top of the \texttt{ARF} classification algorithm applied to the \texttt{Phishing} dataset  We can see the relatively constant cumulative log loss improvement of the protected classification algorithm, which also results in better calibration overall.

\section{Conclusions and further work}

This paper introduced the extension to original work on protected classification reported in \cite{vovk2021protected} through the use of a novel library which can applied to a range of binary and multi-class classification experiments with and without dataset shift. The empirical results suggest that the method is applicable to a broad range of problems and can help to produce more calibrated outputs when applied on top of the underlying classifiers or other equivalent post-hoc calibration methods. Further work could extend this technique to some of the more recent developments generative AI, for example, calibrating probabilistic token outputs of Large Language Models (LLM) in an online setting. We hope that the research community can benefit from the use of this library in as an additional tool in handing dataset shifts in other machine learning problems.

% Acknowledgements and Disclosure of Funding should go at the end, before appendices and references
%\begin{credits}
%\subsubsection{\ackname} I am incredibly grateful to Professors Vladimir Vovk and Alexander Gammerman for valuable discussions during the course of writing this paper.
%\end{credits}

\newpage

\vskip 0.2in
\bibliographystyle{splncs04nat}
\bibliography{arxiv_2025_bbl}

\end{document}